\documentclass[10pt,twocolumn,letterpaper]{article}

\usepackage[usenames,dvipsnames,svgnames,table]{xcolor}

\usepackage{iccv}
\usepackage{times}
\usepackage{epsfig}
\usepackage{graphicx} 
\usepackage{amsmath}
\usepackage{amssymb}
\usepackage{textcomp}
\usepackage{subcaption}
\usepackage{cite}

\usepackage{float}
\usepackage{longtable,booktabs}
\usepackage{algorithm}
\usepackage[noend]{algpseudocode}
\usepackage{multirow}
\usepackage[pagebackref=true,breaklinks=true,letterpaper=true,colorlinks,bookmarks=false]{hyperref}

\usepackage{cuted}
\usepackage{capt-of}

\renewcommand{\vec}[1]{\boldsymbol{#1}}

\DeclareMathOperator*{\argmin}{arg\,min}
\newcommand{\T}{\mathcal{T}}
\newcommand{\F}{\mathcal{F}}
\newcommand{\x}{\vec{x}}
\newcommand{\s}{\vec{s}}

\newcommand{\red}[1]{\textcolor[rgb]{1.00,0.00,0.00}{#1}}
\newcommand{\grn}[1]{\textcolor[rgb]{0.20,0.80,0.20}{#1}}
\newcommand{\blu}[1]{\textcolor[rgb]{0.00,0.00,1.00}{#1}}

\iccvfinalcopy 
\def\iccvPaperID{1105} 

\ificcvfinal\fi
\usepackage{fancyhdr}
\fancypagestyle{plain}{%
	\fancyhf{}
	\chead{\textit{To appear in the 2019 IEEE International Conference on Computer Vision (ICCV)}}
	\cfoot{1}
}
\begin{document}

	\title{CompenNet++: End-to-end Full Projector Compensation}
	
	\author{Bingyao Huang\\
		Temple University\\
		{\tt\small bingyao.huang@temple.edu}
		\and Haibin Ling\thanks{Corresponding author.}\\
		Stony Brook University\\
		{\tt\small hling@cs.stonybrook.edu}
	}
	
	\maketitle
	

	\begin{abstract}
		Full projector compensation aims to modify a projector input image such that it can compensate for both geometric and photometric disturbance of the projection surface. Traditional methods usually solve the two parts separately, although they are known to correlate with each other. In this paper, we propose the first end-to-end solution, named CompenNet++, to solve the two problems jointly. Our work non-trivially extends CompenNet~\cite{huang2019compennet}, which was recently proposed for photometric compensation with promising performance.
		First, we propose a novel geometric correction subnet, which is designed with a cascaded coarse-to-fine structure to learn the sampling grid directly from photometric sampling images. Second, by concatenating the geometric correction subset with CompenNet, CompenNet++ accomplishes full projector compensation and is end-to-end trainable. Third, after training, we significantly simplify both geometric and photometric compensation parts, and hence largely improves the running time efficiency. Moreover, we construct the first setup-independent full compensation benchmark to facilitate the study on this topic. In our thorough experiments, our method shows clear advantages over previous arts with promising compensation quality and meanwhile being practically convenient.
	\end{abstract}
	
	\section{Introduction}\label{sec:intro}
	
	\begin{figure}[!t]
		\begin{center}
			\vspace{-4mm}\includegraphics[width=.98\linewidth]{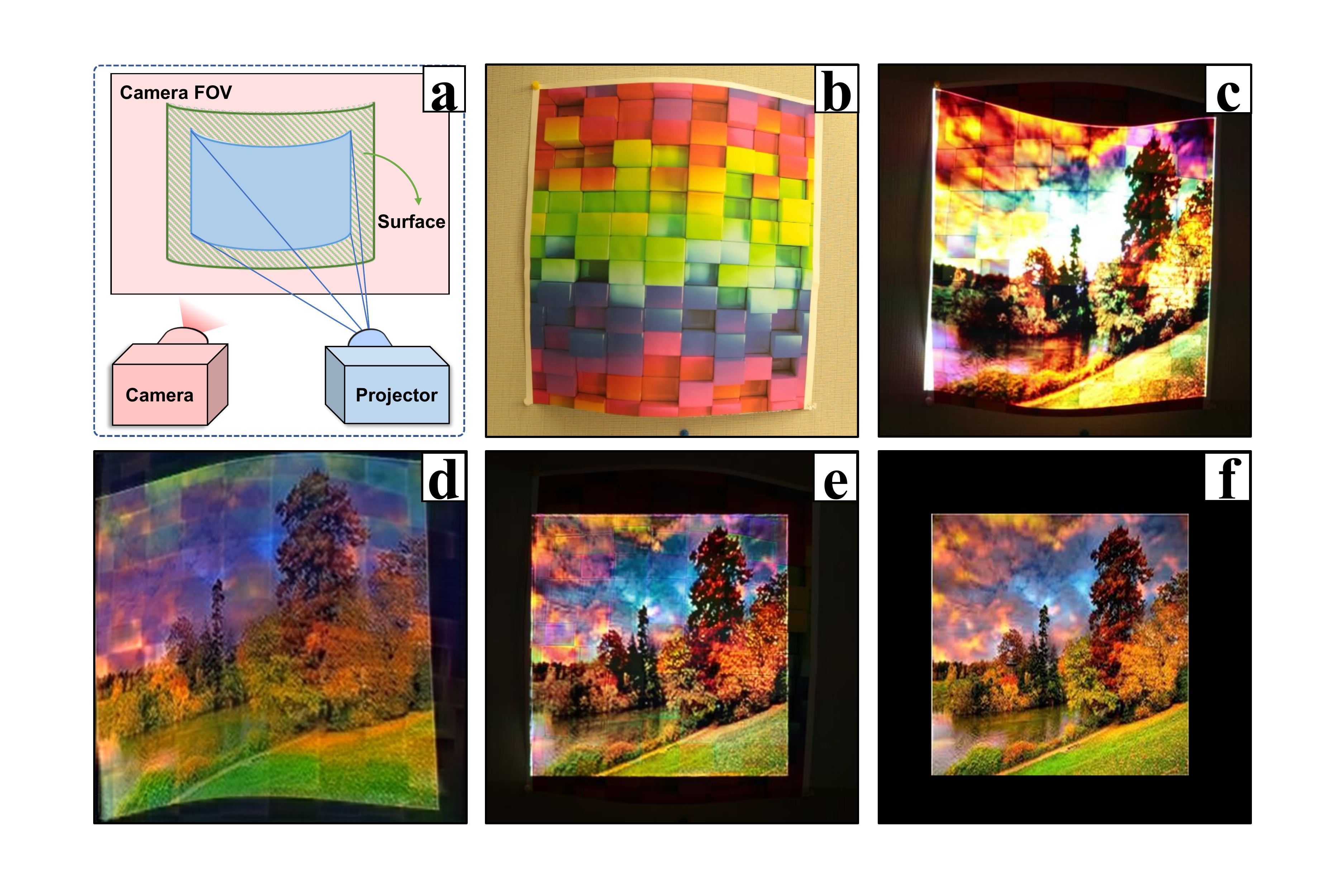}
			\caption{Full projector geometric correction and photometric compensation: \textbf{(a)} system setup with nonplanar and textured surface \textbf{(b)}, \textbf{(c)} projection result without compensation, 
				\textbf{(d)} fully compensated image by our method, \textbf{(e)} camera-captured compensated projection result (\ie (d) projected onto (b)), and \textbf{(f)} desired visual effect. Comparing (c) and (e) we see clearly improved geometry, color and details. }\label{fig:tisser}
		\end{center}
		\vspace{-7mm}
	\end{figure}

	With the recent advance in projector technologies, projectors have been gaining increasing popularity with many applications~\cite{Geng2011,raskar2003ilamps, yoshida2003virtual,grossberg2004making, bimber2005embedded, aliaga2012fast,siegl2015real, siegl2017adaptive, narita2017dynamic, grundhofer2018recent, huang2018single}. Existing systems typically request the projection surface (screen) to be planar, white and textureless, under reasonable environment illumination. These requests often create bottlenecks for generalization of projector systems. Projector geometric correction \cite{raskar2001self, raskar2003ilamps, tardif2003multi, boroomand2016saliency, narita2017dynamic} and photometric compensation~\cite{yoshida2003virtual, ashdown2006robust, aliaga2012fast, grundhofer2015robust, huang2019compennet}, or full projector geometric correction and photometric compensation\footnote{In the rest of the text, we call it \textit{full  compensation} for conciseness.}  \cite{raskar2001shader,bimber2005embedded, harville2006practical, siegl2015real, siegl2017adaptive, asayama2018fabricating} aim to address this issue by modifying a projector input image to compensate for the projection setup geometry and associated photometric environment. An example from our solution is illustrated in Fig.~\ref{fig:tisser}, where the compensated projection result (e) is clearly more visually pleasant than the uncompensated one in (c).
	
	A typical full compensation system consists of a projector-camera (pro-cam) pair and a nonplanar textured projection surface placed at a fixed distance and orientation (Fig.~\ref{fig:tisser}(a)). Most existing methods work in two separate steps: (1) geometric surface modeling, \eg, via a sequence of structured light (SL) patterns \cite{Geng2011, moreno2012simple}, and  (2) color compensation on top of the geometrically corrected projection. 
	Despite relatively easy to implement, this two-step pipeline has two major issues.
	First, geometric mapping/correction is usually performed offline and assumed independent of photometric compensation. This step typically requests certain patterns (\eg SL grid) that may be disturbed by surface appearance (\eg reflection, see Fig.~\ref{fig:compare_existing}). 
	Second, due to the extremely complex photometric process involved in pro-cam systems, it is hard for traditional photometric compensation solutions to faithfully accomplish their task. 
	
	
	Recently, an end-to-end photometric compensation algorithm named CompenNet~\cite{huang2019compennet} is introduced and shows great advantage of deep neural networks over traditional solutions. However, it leaves the geometric correction part untouched and hence is restricted on planar surfaces. Moreover, as will be shown in this paper, its running time efficiency still has room to improve.

	To address the above mentioned issues, in this paper we propose the first end-to-end solution, named \emph{CompenNet++}, for full projector compensation. CompenNet++ non-trivially extends CompenNet and jointly solves both geometric correction and photometric compensation in a unified convolutional neural network (CNN) pipeline. In particular, by taking into consideration of both geometric and photometric ingredients in the compensation formulation, we carefully design CompenNet++ as composed of two subnets. The first subnet is a novel cascaded coarse-to-fine sampling grid prediction subnet, named \textit{WarpingNet} (Fig.~\ref{fig:warpNet}), which performs geometric correction; while the second subnet is an improved version of the original CompenNet for photometric compensation. It is worth highlighting that the two subnets are concatenated directly, which makes CompenNet++ end-to-end trainable.
	
	Moreover, following evaluation procedure in~\cite{huang2019compennet}, we construct the first known setup-independent full compensation evaluation benchmark for nonplanar textured surfaces. The proposed CompenNet++ is evaluated on this benchmark that is carefully designed to cover various challenging factors. In the experiments, CompenNet++ demonstrates clear advantages compared with state-of-the-arts.	
	
	Our contributions can be summarized as follows:
	\vspace{-.7em}
	\begin{enumerate}
		\setlength{\itemsep}{0pt}
		\setlength{\parsep}{1pt}
		\setlength{\parskip}{1pt}
		\item The proposed CompenNet++ is the first end-to-end full compensation system.
		\item Compared with two-step methods, CompenNet++ learns the geometric correction without extra sampling images and outperforms the compared counterparts.
		\item Two task-specific weight initialization approaches are proposed to ensure the convergence and stability of CompenNet++.
		\item Novel simplification techniques are developed to improve the running time efficiency of CompenNet++.
	\end{enumerate}
	\vspace{-2.5mm}The source code, benchmark and experimental results are available at {\small\url{https://github.com/BingyaoHuang/CompenNet-plusplus}}.

	\section{Related Works}\label{sec:related_works}
	In this section, we review existing projector compensation methods in roughly two types: full compensation \cite{raskar2001shader,bimber2005embedded, harville2006practical, siegl2015real, siegl2017adaptive, shahpaski2017simultaneous}  and partial ones \cite{nayar2003projection, grossberg2004making, sajadi2010adict,  grundhofer2015robust,ashdown2006robust,aliaga2012fast, takeda2016inter, li2018practical, huang2019compennet}.
	
	\subsection{Full compensation methods}
	\vspace{-1.4mm}Full compensation methods perform  both geometric correction and photometric compensation. The pioneer work by Raskar \etal~\cite{raskar2001shader} creates projection mapping animations on nonplanar colored objects with two projectors. Despite compensating both geometry and photometry, manual registrations using known markers are required.
	Harville \etal~\cite{harville2006practical} propose a full multi-projector compensation method applied to a white curved screen. The pro-cam pixel correspondences are obtained via 8-12 SL images. Despite being effective to blend multiple projector's color, this method assumes a textureless projection surface.
	
	Recently, Siegl \etal~\cite{siegl2015real, siegl2017adaptive} perform full compensation on nonplanar Lambertian surfaces for dynamic real-time projection mapping. Similar to \cite{harville2006practical}, they assume the target objects are white and textureless. Asayama \etal \cite{asayama2018fabricating} attach visual markers to nonplanar textured surfaces for real-time object pose tracking. To remove the disturbance of the markers, photometric compensation is applied to hide the markers from the viewer, and extra IR cameras/emitters are required accordingly.	Shahpaski \etal \cite{shahpaski2017simultaneous} embed color squares in the projected checkerboard pattern to calibrate both geometry and gamma function. Although only two shots are required, this method needs a pre-calibrated camera and another planar printed checkerboard target. Moreover, it only performs a uniform gamma compensation without compensating the surface, and thus may not work well on nonplanar textured surfaces. 
	
	\subsection{Partial compensation methods}
	\vspace{-1.4mm}Compared to full compensation methods, partial compensation ones typically perform either geometric correction \cite{raskar2001self, raskar2003ilamps, tardif2003multi, boroomand2016saliency, narita2017dynamic} or photometric compensation \cite{yoshida2003virtual, ashdown2006robust, aliaga2012fast, grundhofer2015robust, huang2019compennet}. Due to the strong mutual-dependence between geometric correction and photometric compensation and to avoid propagated errors from the other part, these methods assume the other part is already performed as a prerequisite.
	
	\vspace{.2mm}\noindent\textbf{Geometric correction}. Without using specialized hardware, such as a coaxial pro-cam pair \cite{fujii2005projector}, pro-cam image pairs' geometric mapping need to be estimated using methods such as SL \cite{raskar2001self,raskar2003ilamps, tardif2003multi, boroomand2016saliency}, markers \cite{narita2017dynamic} or homographies \cite{huang2019compennet}. Raskar \etal \cite{raskar2003ilamps} propose a conformal texture mapping method to geometrically register multiple projectors for nonplanar surface projections, using SL and a calibrated camera. Tardif \etal \cite{tardif2003multi} achieve similar results without calibrating the pro-cam pair.  The geometrically corrected image is generated by SL inverse mapping. Similarly, Boroomand \etal \cite{boroomand2016saliency} propose a saliency-guided SL geometric correction method. 
	Narita \etal \cite{narita2017dynamic} use IR ink printed fiducial markers and a high-frame-rate camera for dynamic non-rigid surface projection mapping, which requires extra devices as \cite{asayama2018fabricating}.
	
	\vspace{.2mm}\noindent\textbf{Photometric compensation}. These methods assume the pro-cam image pairs are registered as a prerequisite and can be roughly categorized into two types: context-independent \cite{nayar2003projection, grossberg2004making, sajadi2010adict,  grundhofer2015robust} and context-aware ones \cite{ashdown2006robust,aliaga2012fast, takeda2016inter, li2018practical, huang2019compennet}, where context-aware ones typically assume pro-cam pixels one-to-one mapping and context-aware ones also consider neighborhood/global information. A detailed review can be found in \cite{grundhofer2018recent}.
	Previous compensation methods either assume the compensation is partially done as a prerequisite or perform two-step compensation separately. However, separating the two steps is known to subject to suboptimal solutions. To the best of our knowledge, there exists no previous method that performs simultaneous full pro-cam image geometric correction and projector photometric compensation.
	
	Belonging to the full compensation regime, our CompenNet++ is the first to jointly learn geometric correction and photometric compensation in an end-to-end framework. Though some part of CompenNet++ is based on CompenNet, there are significant differences: (1) CompenNet++ is for full projector compensation; (2) the photometric part in CompenNet++ extends CompenNet by trimming the surface image branch, and hence improves runtime efficiency with no performance drop; and (3) the concatenation of the geometric and photometric parts in CompenNet++ allows both parts to be jointly trained end-to-end. 

	\begin{figure*}[!t]
		\begin{center}
			\vspace{-2mm}\includegraphics[width=1\linewidth]{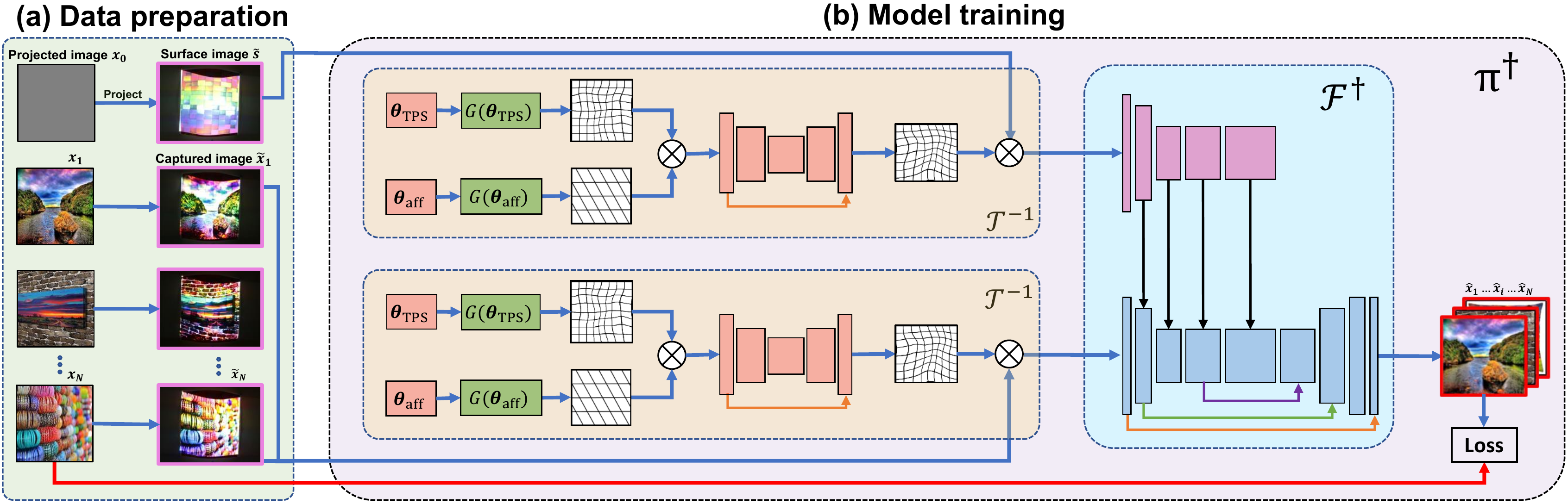}
			
			\caption{Training of CompenNet++ in two major steps. \textbf{(a)} Project and capture a surface image and a set of sampling images. \textbf{(b)} CompenNet++, \ie, {\small $\pi^{\dagger}_{\vec{\theta}}$}, is trained using the data prepared in (a). 
			}   \label{fig:flowchart}
			\vspace{-5mm}		\end{center}
	\end{figure*}
	\section{End-to-end Full Projector Compensation}\label{sec:problem}
	\subsection{Problem formulation}\label{subsec:problem_formulation}
	
	Our full projector compensation system consists of an uncalibrated pro-cam pair and a nonplanar textured projection surface placed at a fixed distance and orientation (Fig.~\ref{fig:tisser}(a)). Following the convention of \cite{huang2019compennet} we extend the photometric compensation formulation to a full compensation one.  Denote a projector input image by $\x$, the  composite  geometric projection and radiometric transfer function by $ \pi_p $ and projector geometric and photometric intrinsics and extrinsics by $ \vec{p} $. Then, the projected radiance can be denoted by $\pi_p(\x, \vec{p})$. Let the composite surface reflectance, geometry and pose be $\s$, surface bidirectional reflectance distribution function (BRDF) be $\pi_s$,  the global lighting irradiance distribution be $\vec{g}$, camera's composite capturing function be $\pi_c$, and its composite intrinsics and extrinsics be $ \vec{c} $. Then the camera-captured image $\tilde{\x}$ is given by\footnote{As in \cite{huang2019compennet}, we use `tilde' ($\tilde{\x}$) to indicate a camera-captured image.}:
	\vspace{-1mm}\begin{equation}\label{eq:ref}
	\vspace{-1mm}\tilde{\x} = \pi_c\big(\pi_s\big(\pi_p( \x , \vec{p} ), \vec{g}, \s\big), \vec{c}\big)
	\end{equation}	
	Note the composite geometric and radiometric process in Eq.~\ref{eq:ref} is very complex and obviously has no  closed form solution. Instead, we find that $\vec{p}$ and $\vec{c}$ are constant once the setup is fixed, thus, we disentangle the geometric and radiometric transformations and absorb $\vec{p}$ and $\vec{c}$ in two functions: {\small$\T:\mathbb{R}^{H_1\times W_1\times 3}\mapsto \mathbb{R}^{H_2\times W_2\times 3}$} that geometrically warps a projector input image to camera-captured image; and {\small$\F:\mathbb{R}^{H_1\times W_1\times3}\mapsto \mathbb{R}^{H_1\times W_1\times3} $} that photometrically transforms a projector input image to an uncompensated camera capture image (aligned with projector's view). Thus, Eq.~\ref{eq:ref} can be reformulated as:
	\vspace{-1mm}\begin{equation}\label{eq:cmp_TF}
	\vspace{-1mm}\tilde{\x} = \T(\F(\x; \vec{g}, \s))
	\end{equation}

	Full projector compensation aims to find a projector input image $ \x^{*} $, named \emph{compensation image} of $ \x $, such that the viewer perceived projection result is the same as the ideal desired viewer perceived image\footnote{In practice, it depends on the optimal displayable area (Fig.~\ref{fig:test_optimal}).}, \ie,
	\vspace{-1mm}\begin{equation}\label{eq:cmp_cam}
	\vspace{-1mm}\T(\F(\x^{*}; \vec{g}, \s))  = \x
	\end{equation}
	Thus the  compensation image $ \x^{*} $ in Eq.~\ref{eq:cmp_cam} is solved by:
	\vspace{-1mm}\begin{equation}\label{eq:x*}
	\vspace{-1mm}
	\x^{*} = \F^{\dagger}(\T^{-1}(\x); \vec{g}, \s).
	\end{equation}
	
	Following \cite{huang2019compennet}, we capture the spectral interactions between $ \vec{g} $ and $ \s $  using a camera-captured surface image $\tilde{\s} $ under the global lighting and the projector backlight:
	\vspace{-1mm}\begin{equation}\label{eq:g_s}
	\vspace{-1mm}\tilde{\s} = \T(\F(\x_0; \vec{g}, \s)),
	\end{equation}
	where $\x_0$ is set to a plain gray image to provide some illumination.
	
	It is worth noting that other than the surface patches illuminated by the projector, the rest part of the surface outside the projector FOV does not provide useful information for compensation (Fig.~\ref{fig:tisser}(a) green part), thus $\tilde{\s}$ in Eq.~\ref{eq:g_s} can be approximated by a subregion of camera-captured image  ${\small \T^{-1}(\tilde{\s})}$ (Fig.~\ref{fig:tisser}(a) blue part).	Substituting $\vec{g}$ and $\s$ in Eq.~\ref{eq:x*} with ${\small \T^{-1}(\tilde{\s})}$ , we have the compensation problem as
	\begin{equation}\label{eq:pi_dagger}
	\x^{*} = \F^{\dagger}\big(\T^{-1}(\x); \T^{-1}(\tilde{\s})\big),
	\end{equation}
	where $ \F^{\dagger} $ is the pseudo-inverse of $ \F $ and $ \T^{-1} $ is the inverse of the geometric transformation $ \T $. Obviously, Eq.~\ref{eq:pi_dagger} has no closed form solution.

	\subsection{Learning-based formulation}
	\vspace{-1mm}Investigating the formulation in \S\ref{subsec:problem_formulation} we find that:
	\vspace{-1mm}\begin{equation}\label{eq:final}
	\vspace{-1mm}\tilde{\x} = \T\big(\F(\x; \s)\big) \ \ \Rightarrow \ \ \x = \F^{\dagger}\big(\T^{-1}(\tilde{\x}); \T^{-1}(\tilde{\s})\big)
	\end{equation}
	We model $\F^{\dagger}$ and $ \T^{-1} $ jointly with a deep neural network named \emph{CompenNet++} and denoted as $\pi^{\dagger}_{\vec{\theta}}$ (Fig.~\ref{fig:flowchart}(b)):
	\vspace{-1mm}\begin{equation}\label{eq:cnn_pred}
	\vspace{-1mm}
	\vec{\hat{x}} = \pi^{\dagger}_{\vec{\theta}}(\tilde{\x}; \tilde{\s}),
	\end{equation}
	where $ \vec{\hat{x}} $ is the compensation of $\tilde{\x}$ (not $\x$) and $ \vec{\theta}= \{\vec{\theta}_\F ,\vec{\theta}_\T\} $ contains the learnable network parameters. In the rest of the paper, we abuse the notation {\small$ \pi^{\dagger}_{\vec{\theta}}(\cdot, \cdot) \equiv \F^{\dagger}_{\vec{\theta}_\F}\big(\T^{-1}_{\vec{\theta}_\T}(\cdot); \T^{-1}_{\vec{\theta}_\T}(\cdot)\big)$} for conciseness. Note that {\small $ \F^{\dagger} $} rather than $ \pi^{\dagger} $ here is the equivalent $ \pi^{\dagger} $ in  \cite{huang2019compennet}.	
	
	We train CompenNet++ over sampled image pairs like $(\tilde{\x}, \x)$ and a surface image $\tilde{\s}$ (Fig.~\ref{fig:flowchart}(a)). By using Eq.~\ref{eq:cnn_pred}, we can generate a set of $N$ training pairs, denoted as $\mathcal{X}=\{(\tilde{\x}_i, \x_i)\}_{i=1}^N $. Then, with a loss function $ \mathcal{L}$, CompenNet++ can be learned by
	\vspace{-2mm}\begin{equation}\label{eq:objective}
	\vspace{-3mm}\vec{\theta} = \argmin_{\vec{\theta}'}\sum_i\mathcal{L}\big(\vec{\hat{x}}_i=\pi^{\dagger}_{\vec{\theta}'}(\tilde{\x}_i; \tilde{\s}), \ \x_i\big)
	\end{equation}
	We use the  loss function below to jointly optimize the color fidelity (pixel-wise $ \ell_1 $) and structural similarity (SSIM):
	\vspace{-2mm}\begin{equation}\label{eq:loss}
	\vspace{-2mm}\mathcal{L}  = \mathcal{L}_{\ell_1} + \mathcal{L}_{\text{SSIM}}
	\end{equation}
	The advantages of this loss function are shown in \cite{zhao2017loss, huang2019compennet}.
	
	\subsection{Network design}\label{subsec:cnn_architecture}
	\begin{figure*}[!t]
		\begin{center}
			\includegraphics[width=1.0\linewidth]{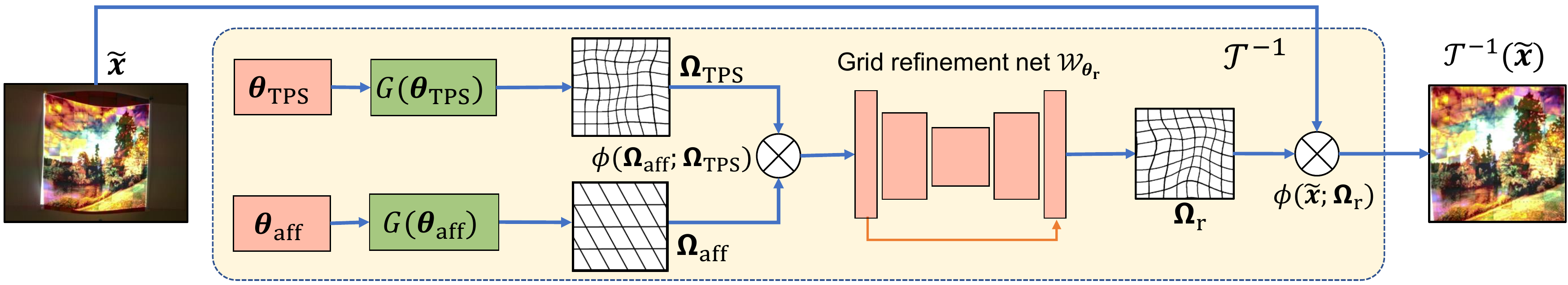}
			\caption{WarpingNet ({\small$ \T^{-1} $}) architecture (activations layers \cite{nair2010rectified, maas2013rectifier} omitted). It warps the input camera-captured image $ \tilde{\x} $  to the projector's view using a cascaded coarse-to-fine structure. The red and green blocks are learnable parameters and grid generation functions, respectively. Operator $ \otimes $ denotes bilinear interpolation, \ie, $ \phi(\cdot; \cdot) $. The grid refinement network {\small$\mathcal{W}_{\vec{\theta}\textsubscript{r}}$} consists of a UNet-like \cite{ronneberger2015u} structure, it generates a refined sampling grid that samples the input image directly.  }\label{fig:warpNet}
		\end{center}
		\vspace{-6mm}
	\end{figure*}
	
	Based on the above formulation, our CompenNet++ is designed with two subnets, a \textbf{WarpingNet} {\small$ \T^{-1} $} that corrects the geometric distortions and warps camera-captured uncompensated images to projector image space; and a \textbf{CompenNet} {\small$ \F^{\dagger} $} that photometrically compensates warped images. The network architecture is shown in Fig.~\ref{fig:flowchart}.  For compactness, we move the detailed parameters of CompenNet++ to the supplementary material.
	
	\vspace{1mm}\noindent\textbf{WarpingNet.} 
	Note directly estimating nonparametric geometric correction is difficult and computationally expensive. Instead, we model the geometric correction as a cascaded coarse-to-fine process, as inspired by the work in~\cite{jaderberg2015spatial,rocco2017convolutional}. As shown in Fig.~\ref{fig:warpNet}, WarpingNet consists of three learnable modules ($ \vec{\theta}\textsubscript{aff}$, $\vec{\theta}$\textsubscript{TPS} and {\small$\mathcal{W}_{\vec{\theta}\textsubscript{r}}$)}, a grid generation function $ G $, a bilinear interpolation-based image sampler $ \phi $,  and three generated sampling grids with increased granularity, ranked as {\small
		$\vec{\Omega}\textsubscript{r}=G(\vec{\theta}\textsubscript{r})>
		\vec{\Omega}\textsubscript{TPS}=G(\vec{\theta}\textsubscript{TPS})>
		\vec{\Omega}\textsubscript{aff}=G(\vec{\theta}\textsubscript{aff})$}.
	
	Specifically, 	$ \vec{\theta}\textsubscript{aff}$ is a 2$\times$3 learnable affine matrix and it warps the input image $ \tilde{\x} $ to approximate projector's front view. Similarly, $ \vec{\theta}\textsubscript{TPS}$ contains (6$\times$6+2)$\times$2 =76 learnable thin plate spline (TPS) \cite{donato2002approximate} parameters and it
	further nonlinearly warps the output of the affine transformed image $ \phi(\tilde{\x}; \vec{\Omega}\textsubscript{aff}) $ to exact projector's view. Unlike \cite{jaderberg2015spatial, rocco2017convolutional}, $ \vec{\theta}\textsubscript{aff}$ and $ \vec{\theta}\textsubscript{TPS}$ are directly learned without using a regression network, which is more efficient and accurate in our case.
	
	Although TPS can approximate nonlinear smooth geometric transformations, its accuracy depends on the number of control points and the spline assumptions. Thus, it may not precisely model image deformations involved in pro-cam imaging process. To solve this issue, we design a grid refinement CNN, \ie, $ \mathcal{W}_{\vec{\theta}\textsubscript{r}} $  to refine the TPS sampling grid. Basically, this net learns a fine displacement for each 2D coordinate in the TPS sampling grid with a residual connection \cite{he2016deep}, giving the refined sampling grid $ \vec{\Omega}\textsubscript{r}$ higher precision. The advantages of our CompenNet++ over a degraded CompenNet++ without grid refinement net (named CompenNet++ w/o refine) are evidenced in Tab.~\ref{tab:compare} and Fig.~\ref{fig:compare_existing}.
	
	Besides the novel cascaded coarse-to-fine structure with a grid refinement network, we propose a novel sampling strategy that improves WarpingNet efficiency and accuracy. Intuitively, the cascaded coarse-to-fine sampling method should sequentially sample the input $ \tilde{\x} $ as
	\vspace{-1mm}\begin{equation}\label{eq:img_sampling}
	\vspace{-1mm}\hspace{-1.mm} \T^{-1}\hspace{-.5mm}(\tilde{\x}) \hspace{-.5mm}=\hspace{-.5mm}  \phi\big(\phi(\phi(\tilde{\x}; \vec{\Omega}\textsubscript{aff}); \vec{\Omega}\textsubscript{TPS});\vec{\Omega}\textsubscript{r} \!\!=\!\!\mathcal{W}_{\vec{\theta}\textsubscript{r}}(\vec{\Omega}\textsubscript{TPS})\big)
	\end{equation}	
	However, the three bilinear interpolations above are not only computationally inefficient but also produce a blurred image. Instead, we perform the sampling in 2D coordinate space, \ie, let the finer TPS grid sample the coarser affine grid, then  refine the grid using $ \mathcal{W}_{\vec{\theta}\textsubscript{r}} $,  as shown in Fig.~\ref{fig:warpNet}. Thus, the output image is given by:
	\vspace{-1mm}
	\begin{equation}\label{eq:grid_sampling}
	\vspace{-1mm}
	\T^{-1}(\tilde{\x}) =	\phi\big(\tilde{\x}; \mathcal{W}_{\vec{\theta}\textsubscript{r}}(\phi(\vec{\Omega}\textsubscript{aff}; \vec{\Omega}\textsubscript{TPS})\big)
	\end{equation}
	This strategy brings two benefits: (1) only two sampling operations are required and thus is more efficient; and (2) since the image sampling is only performed once on $ \tilde{\x} $, the warped image is sharper compared with using Eq.~\ref{eq:img_sampling}.
	
\begin{figure*}[t]
		\begin{center}
			\vspace{-4mm}\includegraphics[width=1.0\linewidth]{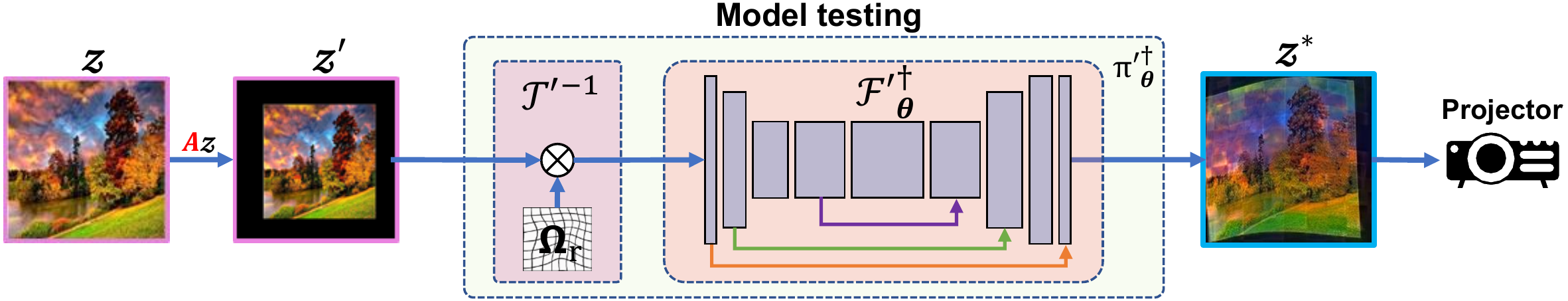}
			\caption{The testing phase of the proposed CompenNet++. Due to our novel WarpingNet structure and sampling strategy, the network is simplified to improve computational and memory efficiency. As we can see the compensation image $ \vec{z}^{*}  $ is both geometrically and photometrically compensated, such that after projection it cancels the geometric and photometric distortions and produce an image that is close to $\vec{z}'$, \ie Fig.~\ref{fig:tisser}(e).}\label{fig:testing}
		\end{center}
		\vspace{-4mm}
	\end{figure*}

	Another novelty of WarpingNet is network simplification owing to the sampling strategy above. During testing, WarpingNet is simplified essentially to a single sampling grid $\vec{\Omega}\textsubscript{r} $, and geometric correction becomes a single bilinear interpolation {\small $ \T^{-1}(\tilde{\x}) =  \phi(\tilde{\x};\vec{\Omega}\textsubscript{r}) $} bringing improved testing efficiency (see Fig.~\ref{fig:testing}).  
	
	
	\vspace{1mm}\noindent\textbf{CompenNet.}
	During training,  {\small$\F^{\dagger}$} takes two WarpingNet transformed images as inputs, \ie, a surface image {\small$ \T^{-1}(\tilde{\s}) $} and a camera-captured image {\small$ \T^{-1}(\tilde{\x}) $}.  The architecture basically follows  \cite{huang2019compennet}, but with two improvements below.
	
	The CompenNet in \cite{huang2019compennet} cannot be directly applied to our CompenNet++ with its original initialization technique, since the joint geometric and photometric process is too complex to learn. Tackling this issue, we propose some useful training techniques in \S\ref{subsec:cnn_training}.
	
	Another improvement is that, for the testing phase, the surface feature autoencoder subset is trimmed by merging into the main backbone as biases (Fig.~\ref{fig:testing}). This network simplification, together with the one on WarpingNet, largely improves the running time and memory efficiency of CompenNet++, without any sacrifice in performance quality.
	
	\begin{figure}[!t]
		\begin{center}
			\includegraphics[width=1\linewidth]{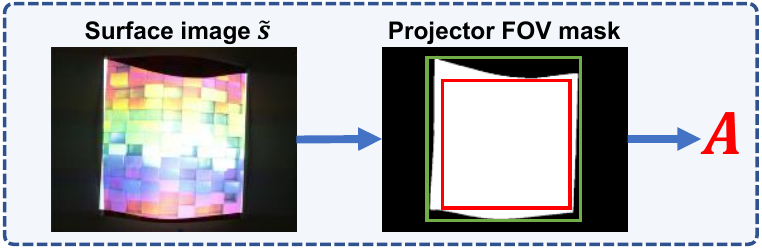}
			\vspace{-4mm}
			\caption{Projector FOV mask, bounding rectangle (green) and optimal displayable area (red). The optimal displayable area is defined as the maximum inscribed rectangle (keep aspect ratio) \cite{raskar2003ilamps}. The affine matrix $ \vec{A} $ is estimated given the displayable area and projector input image size.}
			\label{fig:test_optimal}
		\end{center}
		\vspace{-8mm}
	\end{figure}
	
	\subsection{Training details}\label{subsec:cnn_training}
	Compared with CompenNet \cite{huang2019compennet} training, simultaneously optimizing WarpingNet parameters $ \vec{\theta}_\T$ and CompenNet parameters $\vec{\theta}_\F $ is hard without proper weights initialization and automatic data preprocessing. 
	
	\vspace{.5mm}\noindent\textbf{Projector FOV mask.} 
	According to Eq.~\ref{eq:pi_dagger}, full projector compensation's region of interest is the projector FOV, \ie Fig.~\ref{fig:tisser}(a) blue part. Thus we can compute a projector FOV mask by automatically thresholding the camera-captured surface images with Otsu's method \cite{otsu1979threshold} followed by some morphological operations (Fig.~\ref{fig:test_optimal}). This mask brings threefold benefits: \textbf{(1)} masking out the pixels outside of FOV improves training stability and efficiency; \textbf{(2)} the projector FOV mask is the key to initialize WarpingNet affine weights below and \textbf{(3)} to find the optimal displayable area in \S\ref{subsec:pipeline}. 
	
	\vspace{.5mm}\noindent\textbf{WarpingNet weights initialization.}
	We further improve the training efficiency by providing a task specific prior, \eg, the coarse affine warping branch in WarpingNet aims to transform the input image $ \tilde{\x} $ to projector's front view, as mentioned in \S\ref{subsec:cnn_architecture}. Thus, we initialize the affine parameters $\vec{\theta}$\textsubscript{aff} such that the projector FOV mask's bounding rectangle (Fig.~\ref{fig:test_optimal} green rectangle) is stretched to fill the warped image. Then, $\vec{\theta}$\textsubscript{TPS} and grid refinement net $\mathcal{W}_{\vec{\theta}\textsubscript{r}}$ are initialized with small random numbers at a scale of $10^{-4}$, such that they generate identity mapping.	These task specific initialization techniques provide a reasonably good starting point, allowing CompenNet++ to converge stably and efficiently.
	
	\vspace{.5mm}\noindent\textbf{CompenNet weights initialization.} In \cite{huang2019compennet}, the CompenNet weights are randomly initialized with He's method \cite{he2015delving} and it works well when input images are registered to projector's view offline. In our end-to-end full compensation pipeline, despite with the training techniques above, joint training WarpingNet and CompenNet may subject to suboptimal solutions, \eg, the output images become plain gray. Similar to WarpingNet weights initialization, we introduce some photometric prior knowledge to improve CompenNet stability and efficiency. Inspired by traditional context-independent linear method \cite{nayar2003projection}, we initialize CompenNet to a simple linear channel-independent model such that:
	\vspace{-1mm}\begin{equation}\label{eq:init}
	\vspace{-1mm}\vec{\theta}_{\F} = \argmin_{\vec{\theta}'_{\F}}\sum_i\mathcal{L}\big(\F^{\dagger}_{\vec{\theta}'_{\F}}(\x_i; \dot{\s}), \ \max(0, \x_i - \dot{\s})\big),
	\end{equation}
	where $ \x_i $ is a projector input image and $ \dot{\s} $ is a colorful textured image that mimics the warped surface image {\small $ \T^{-1}(\tilde{\s}) $}. Compared with CompenNet's pre-train method \cite{huang2019compennet}, our approach creates a simple yet effective initialization without any actual projection/capture.
	Note this weight initialization is only performed once and independent of setups. For a new setup, {\small $\vec{\theta}_{\F}$} is initialized by loading the saved weights.
	
	\subsection{Network Simplification}\label{subsec:cnn_testing}
	During testing, the structure of CompenNet++ shown in Fig.~\ref{fig:testing} is simplified from training structure (Fig.~\ref{fig:flowchart}). \textbf{(a)} As mentioned in \S\ref{subsec:cnn_architecture}, due to our novel cascaded coarse-to-fine network design and sampling strategy, WarpingNet can be substituted by a sampling grid and an image sampler shown as $ \T'^{-1} $ in Fig.~\ref{fig:testing}. \textbf{(b)} Similarly, CompenNet's surface feature extraction branch's (the top subnet of $ \F^{\dagger} $) weights and input are both fixed during testing, thus, it is trimmed and replaced by biases to reduce computation and memory usage. The biases are then directly added to the  CompenNet backbone, we denote this simplified CompenNet++ as {\small$ \pi'^{\dagger}_{\vec{\theta}}$}. The two novel network simplification techniques make the proposed CompenNet++ both computationally and memory efficient with no performance drop.
	
	\subsection{Compensation pipeline}\label{subsec:pipeline}
	To summarize, our full projector compensation pipeline consists of three major steps (Fig.~\ref{fig:flowchart} and Fig.~\ref{fig:testing}).  \textbf{(1)} We start by projecting a plain gray image $ \x_0 $,  and $ N $ sampling images $ \x_1, \dots, \x_N $ to the projection surface and capture them using the camera, and denote the captured images as $ \tilde{\s} $ and $ \tilde{\x}_i $, respectively. \textbf{(2)} We gather the $ N $ image pairs $ (\tilde{\x}_i, \x_i) $ and $ \tilde{\s} $ to train the compensation model {\small$ \pi^{\dagger}_{\vec{\theta}} = \{\F^{\dagger}_{\vec{\theta}}, \T^{-1}_{\vec{\theta}}\} $} end-to-end. \textbf{(3)} As shown in Fig.~\ref{fig:testing}, we simplify the trained CompenNet++ to {\small$ \pi'^{\dagger}_{\vec{\theta}}$} using techniques in \S\ref{subsec:cnn_testing}. Finally, for an ideal desired viewer perceived image  $ \vec{z}$, we generate its compensation image $ \vec{z}^{*} $ and project $ \vec{z}^{*} $ to the surface.

	In practice, $ \vec{z} $ is restricted to the surface displayable area. Similar to \cite{raskar2003ilamps}, we find an optimal desired image $ \vec{z}' = \vec{A}\vec{z} $, where $ \vec{A} $ is a 2D affine transformation that uniformly scales and translates the ideal perceived image $ \vec{z} $ to optimally fit the projector FOV as shown in Fig.~\ref{fig:test_optimal} and Fig.~\ref{fig:testing}.

	\subsection{System configuration and implementation.}
	Our projector compensation system consists of a Canon 6D camera and a ViewSonic PJD7828HDL DLP projector with resolutions set to 640$\times$480 and 800$\times$600, respectively. In addition, an Elgato Cam Link 4K video capture card is connected to the camera to improve frame capturing efficiency (about 360ms per frame).
	
	The distance between the camera and the projector is varied in the range of 500mm to 1,000mm and the projection surface is around 1,000mm in front of the pro-cam pair. The camera exposure, focus and white balance modes are set to manual, the global lighting is varied for each setup but fixed during each setup's data capturing and system testing.
	
	CompenNet++ is implemented using PyTorch \cite{paszke2017automatic} and trained using Adam optimizer \cite{kinga2015method} with a penalty factor of $ 10^{-4} $. The initial learning rate is set to $ 10^{-3} $ and decayed by a factor of 5 at the 1,000\textsuperscript{th} iteration. The model weights are initialized using the techniques in \S\ref{subsec:cnn_training}. We train the model for 1,500 iterations on three Nvidia GeForce 1080Ti GPUs with a batch size of 48, and it takes about 15min to finish.
	
	\subsection{Dataset and evaluation protocol}	
	Following \cite{huang2019compennet}, we prepare 700 colorful textured images and use $ N=500 $ for each training set $\mathcal{X}_k$ and $ M = 200$ for each validation  set $\mathcal{Y}_k$. In total $K = 20$ different setups are prepared for training and evaluation, each setup has a nonplanar textured surface. 
	
	We collect the setup-independent validation set of $M$ samples as $\mathcal{Y}=\{(\tilde{\vec{y}}_i, \vec{y}_i)\}_{i=1}^M$, under the same system setup as the training set $\mathcal{X}$. Then the algorithm performance is measured by averaging over similarities between each validation input image $\vec{y}_i$ and its algorithm output $\hat{\vec{y}}_i = \pi^{\dagger}_{\vec{\theta}}(\tilde{\vec{y}}_i; \tilde{\s})$ and reported in Tab.~\ref{tab:compare}. Note we use the same evaluation metrics PSNR, RMSE and SSIM as in \cite{huang2019compennet}.

	\section{Experimental Evaluations}\label{sec:experiments}
	
	\subsection{Comparison with state-of-the-arts}\label{subsec:comparison_existing}
	We compare the proposed full compensation method (\ie  CompenNet++) with four two-step baselines, a context-independent TPS\footnote{Not geometric correction \cite{donato2002approximate}, instead using TPS to model pixel-wise \textit{photometric transfer function}.} model \cite{grundhofer2015robust}, an improved TPS model (explained below), a Pix2pix \cite{isola2017image} model and a CompenNet \cite{huang2019compennet} model on the evaluation benchmark.
	
	\begin{table*}
		\begin{center}
			\caption{Quantitative comparison of compensation algorithms. Results are averaged over $K= 20$ different setups.  The top-3 results of each column in each \textbf{\#Train} section are highlighted as \red{red}, \grn{green} and \blu{blue}, respectively. Note the metrics for uncompensated images are PSNR=9.5973, RMSE=0.5765 and SSIM=0.0767. The metrics for the original TPS \cite{grundhofer2015robust} w/ SL (\#Train=125) are PSNR=16.7271, RMSE= 0.2549 and SSIM=0.5207. 
			}\label{tab:compare}
			\vspace{-3mm}{\footnotesize 
				\begin{tabular}{@{\hspace{0.mm}}l|c@{\hspace{2mm}}c@{\hspace{2mm}}c@{\hspace{2mm}}|c@{\hspace{2mm}}c@{\hspace{2mm}}l|c@{\hspace{2mm}}c@{\hspace{2mm}}l|c@{\hspace{2mm}}c@{\hspace{2mm}}c@{\hspace{0.mm}}}
					\hline
					\multicolumn{1}{c|}{\multirow{2}{*}{\textbf{Model}}} & \multicolumn{3}{c|}{\textbf{\#Train=48}}                & \multicolumn{3}{c|}{\textbf{\#Train=125}}               & \multicolumn{3}{c|}{\textbf{\#Train=250}}               & \multicolumn{3}{c}{\textbf{\#Train=500}}               \\
					\multicolumn{1}{c|}{}                                & \textbf{PSNR}$\uparrow$ & \textbf{RMSE}$\downarrow$ & \textbf{SSIM}$\uparrow$ & \textbf{PSNR}$\uparrow$ & \textbf{RMSE}$\downarrow$ & \textbf{SSIM}$\uparrow$ & \textbf{PSNR}$\uparrow$ & \textbf{RMSE}$\downarrow$ & \textbf{SSIM}$\uparrow$ & \textbf{PSNR}$\uparrow$ & \textbf{RMSE}$\downarrow$ & \textbf{SSIM}$\uparrow$ \\\hline
					TPS \cite{grundhofer2015robust} textured w/ SL                                  & 18.0297       & 0.2199        & 0.5390        & 18.0132       & 0.2205        & 0.5687        & 18.0080       & 0.2206        & 0.5787        & 17.9746       & 0.2215        & 0.5830        \\
					Pix2pix \cite{isola2017image} w/ SL                                       & 17.7160       & 0.2271        & 0.5068        & 17.1141       & 0.2468        & 0.5592        & 16.5236       & 0.2669        & 0.5763        & 19.4160       & 0.1903        & 0.6196        \\				
					CompenNet \cite{huang2019compennet} w/ SL                                     & \red{20.2023}       & \red{0.1722}        & \grn{0.6690}        & \grn{20.7684}       & \grn{0.1609}        & \grn{0.7022}        & \grn{20.8347}       & \grn{0.1596}        & \grn{0.7142}        & \grn{20.9552}       & \grn{0.1573}        & \grn{0.7117}        \\
					CompenNet++ w/o refine                              & 19.4139       & 0.1909        & 0.6252        & \blu{20.6061}       & \blu{0.1635}        & 0.6958        & \blu{20.7307}       & \blu{0.1613}        & \blu{0.7106}        & \blu{20.9172}       & \blu{0.1577}        & \blu{0.7113}        \\
					CompenNet++                                         & \blu{19.8552}       & \blu{0.1781}        & \blu{0.6637}        & \red{20.7947}       & \red{0.1598}        & \red{0.7116}        & \red{20.8959}       & \red{0.1581}        & \red{0.7227}        & \red{21.1127}       & \red{0.1540}        & \red{0.7269}        \\
					CompenNet++ fast                                    & \grn{19.9696}       & \grn{0.1760}        & \red{0.6699}        & 20.5171 & 0.1650 & \blu{0.7001}        & 20.5795       & 0.1638        & 0.7063        & 20.6711       & 0.1622        & 0.7081        \\
					CompenNet++ faster                                  & 19.2536       & 0.1912        & 0.6249        & 19.5309       & 0.1844        & 0.6546        & 19.7212       & 0.1806        & 0.6613        & 19.6989       & 0.1811        & 0.6574        \\ \hline
				\end{tabular}
			}		
		\end{center}
		\vspace{-3mm}
	\end{table*}

	\begin{figure*}[!t]
		\begin{center}
			\includegraphics[width=1\linewidth]{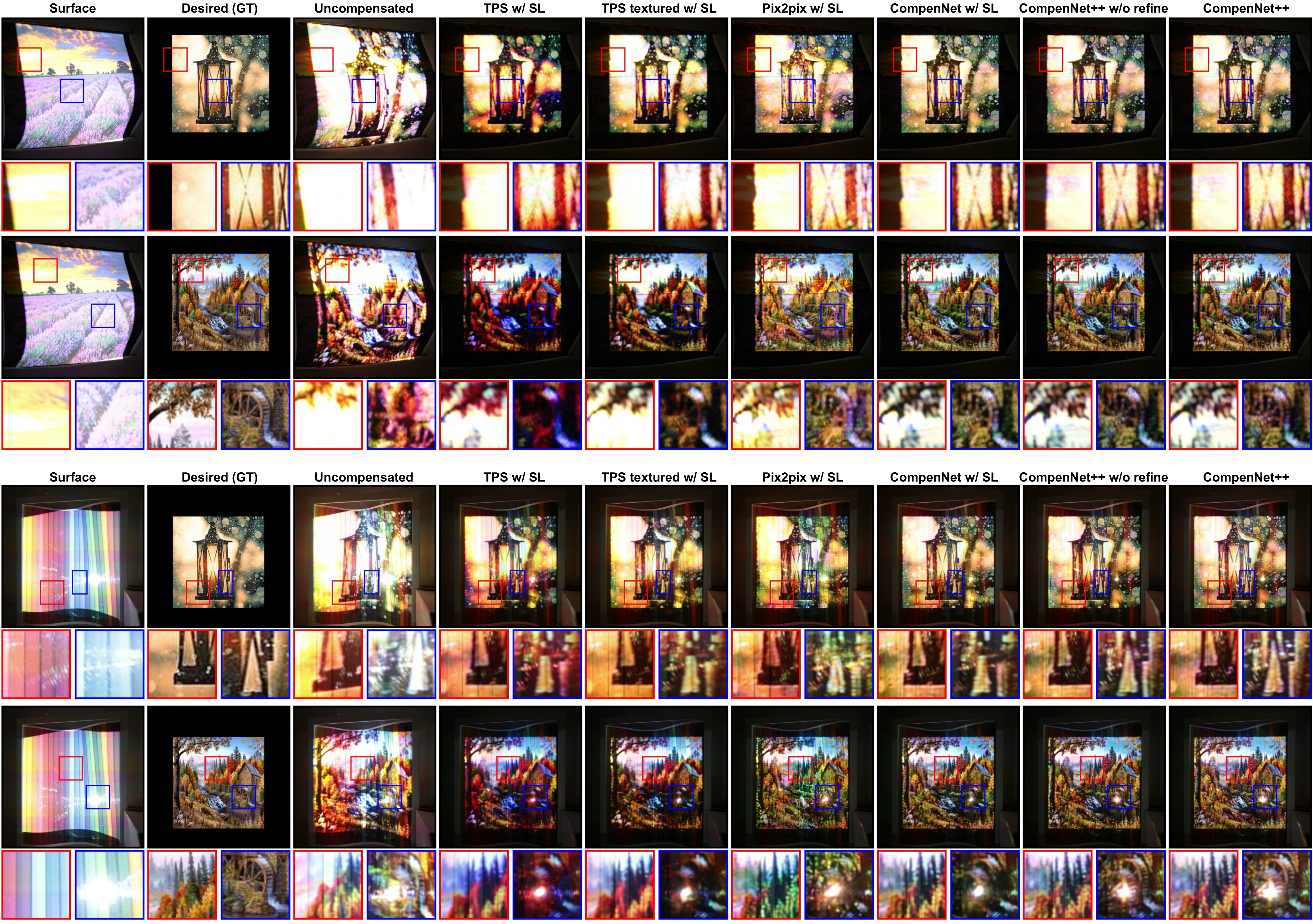}
			\vspace{-5mm}
			\caption{Qualitative comparison of TPS \cite{ grundhofer2015robust} w/ SL, TPS textured w/ SL,  Pix2pix \cite{isola2017image} w/ SL, CompenNet \cite{huang2019compennet} w/ SL, proposed CompenNet++ w/o refine and proposed CompenNet++ on two different surfaces. The 1\textsuperscript{st} to 3\textsuperscript{rd} columns are the camera-captured projection surface, desired viewer perceived image and camera-captured uncompensated projection, respectively.	The rest columns are the compensation results of different methods. Each image is provided with two zoomed-in patches for detailed comparison. More comparisons are provided in supplementary materials.
				\label{fig:compare_existing}
			}
		\end{center}
		\vspace{-5mm}
	\end{figure*}

	To fairly compare two-step methods, we use the same SL warping for geometric correction. We first project 42 SL patterns to establish pro-cam pixel-to-pixel mapping using the approach in \cite{moreno2012simple}, the mapping coordinates are then bilinear-interpolated to fill missing correspondences. Afterwards, we capture 125 pairs of plain color sampling image as used in the original TPS method~\cite{grundhofer2015robust} for photometric compensation, we warp the sampling image to projector's view using SL and name this  method \textbf{TPS w/ SL}. We also fit the TPS method using SL-warped diverse textured training set $\mathcal{X}_k$, and name this method \textbf{TPS textured w/ SL}.
	
	The experiment results in Tab.~\ref{tab:compare} and Fig.~\ref{fig:compare_existing} show clear improvement of TPS textured over the original TPS method. Our explanations are \textbf{(a)} compared with plain color images, the textured training images and validation/testing images share a more similar distribution. \textbf{(b)} Although original TPS method uses $ 5^3 $ plain color images, each projector pixel's R/G/B channel only has five different intensity levels, training the TPS model using these samples may lead to a suboptimal solution. While our colorful textured samples evenly cover the RGB space at each projector pixel, resulting a more faithful sampling of the photometric transfer function.
	
	To demonstrate the difficulty of full compensation problem, we compare with a deep learning-based image-to-image translation model Pix2pix\footnote{{\scriptsize \url{https://github.com/junyanz/pytorch-CycleGAN-and-Pix2pix}}} \cite{isola2017image} trained on the same SL-warped $\mathcal{X}_k$ as TPS textured w/ SL, we name it \textbf{Pix2pix w/ SL}. We use the same adaptation as \cite{huang2019compennet}, except that Pix2pix is trained for 12,000 iterations to match the training time of our model. The results show that the proposed CompenNet++ outperforms Pix2pix w/ SL, demonstrating that the full compensation problem cannot be well solved by a general deep-learning based image-to-image translation model.
	
	We then compare our method with the partial compensation model CompenNet \cite{huang2019compennet}, we train it with the same SL-warped training set $\mathcal{X}_k$ as TPS textured w/ SL and Pix2pix w/ SL, and name this two-step method \textbf{CompenNet w/ SL}. The quantitative and qualitative comparisons are shown in Tab.~\ref{tab:compare} and Fig.~\ref{fig:compare_existing}, respectively.
	
	\subsection{Effectiveness of the proposed CompenNet++}\label{subsec:compare_warp_net}
	Tab.~\ref{tab:compare} clearly shows that CompenNet++ outperforms other two-step methods. 
	This indicates that \textbf{(a)} even without building pixel-to-pixel mapping using SL, the geometry correction can be learned directly from the photometric sampling images. \textbf{(b)} Solving full compensation problem separately may lead to suboptimal solution and the two steps should be solved jointly, as proposed by CompenNet++. \textbf{(c)} Besides outperforming CompenNet w/ SL, we use 42 less images than two-step SL-based method. 
	
	We explain why two-step methods may find suboptimal solution in Fig.~\ref{fig:compare_existing}, where SL decoding errors affect the photometric compensation accuracy. As shown in the 1\textsuperscript{st} row red zoomed-in patches, compared with end-to-end methods (last two columns), SL-based two-step methods (4\textsuperscript{th}-7\textsuperscript{th} columns) produce curved edges, due to inaccurate SL warping. Furthermore,
	in the	3\textsuperscript{rd} and 4\textsuperscript{th} rows, the nonplanar surface is behind a glass with challenging specular reflection. Comparing the two groups, specifically the blue zoomed-in patches, we see unfaithful compensations by the SL-based two-step methods, whereas, end-to-end methods \textbf{CompenNet++ w/o refine} and \textbf{CompenNet++} show finer geometry, color and details. This is because SL suffers from decoding errors due to specular reflection and creates false mappings, then the mapping errors propagate to the photometric compensation stage.  This issue is better addressed by the proposed CompenNet++, where global geometry and photometry information is considered in full compensation. In summary, CompenNet++ not only brings improved performance than two-step SL-based methods, but also waives 42 extra SL projections/captures, and meanwhile being insensitive to specular highlights.

	To demonstrate the practicability of CompenNet++ when efficiency is preferred over quality, \ie, less data and shorter training time, we train CompenNet++ using only 48  images and reduce the training iterations to 1,000/500 and batch size to 24/16, we name the efficient methods \textbf{CompenNet++ fast/faster} and it takes only 5min/2.5min to finish training. As shown in Tab.~\ref{tab:compare}, even when trained with only 48 images, \textbf{CompenNet++ fast/faster} still outperform TPS textured w/ SL and Pix2pix w/ SL trained with 500 images on SSIM.

	\subsection{Effectiveness of the grid refinement network}\label{subsec:compare_grid_refine}
	To demonstrate the effectiveness of the sampling grid refinement network $\mathcal{W}_{\vec{\theta}\textsubscript{r}}$ (Eq.~\ref{eq:grid_sampling} and Fig.~\ref{fig:warpNet}), we create a degraded CompenNet++ by removing $\mathcal{W}_{\vec{\theta}\textsubscript{r}}$, and name it \textbf{CompenNet++ w/o refine}. As reported in Tab.~\ref{tab:compare}, CompenNet++ clearly outperforms this degraded model, showing the effectiveness of  the grid refinement network $\mathcal{W}_{\vec{\theta}\textsubscript{r}}$. 
	

	\begin{figure}[!t]
		\begin{center}
			\includegraphics[width=1\linewidth]{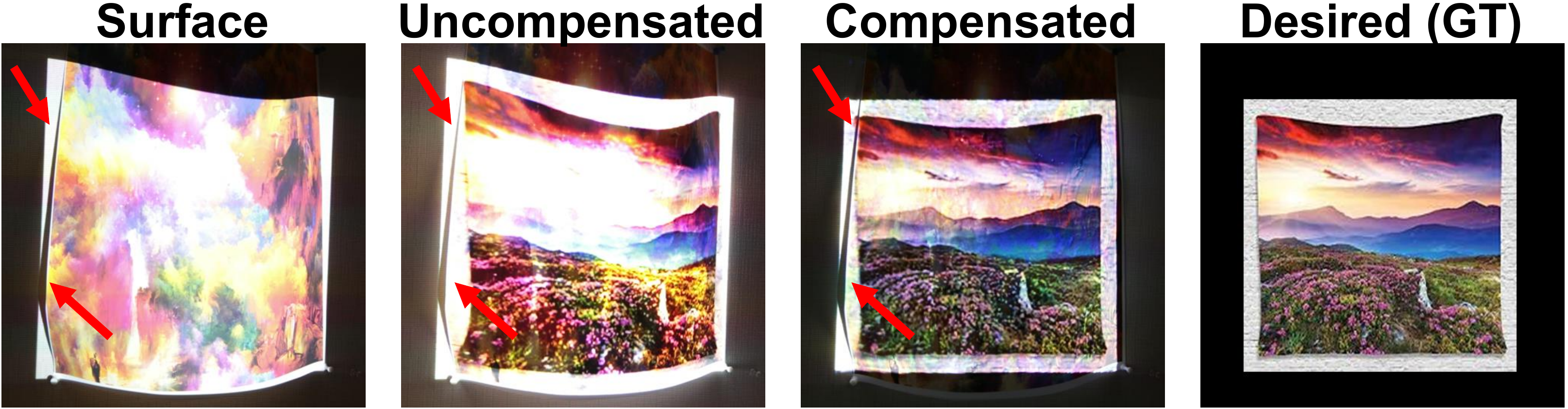}
			\vspace{-4mm}
			\caption{Failed example. CompenNet++ is unable to compensate self-occlusion regions as pointed by red arrows.
				\label{fig:failed}
			}
		\end{center}
		\vspace{-6mm}
	\end{figure}

	\section{Conclusions and Limitations}\label{sec:conclusions}
	In this paper, we extend the partial projector compensation model CompenNet to a full compensation pipeline named CompenNet++. With the novel cascaded coarse-to-fine WarpingNet, task specific training and efficient testing strategies, CompenNet++ provides the first end-to-end simultaneous projector geometric correction and photometric compensation. The effectiveness of our formulation and architecture is verified by comprehensive evaluations. The results show that our end-to-end full compensation outperforms state-of-the-art two-step methods both qualitatively and quantitatively. 
	
	\noindent\textbf{Limitations.} We assume each single patch of the projection surface can be illuminated by the projector. That said, CompenNet++ may not work well on complex surfaces with self-occlusion (Fig.~\ref{fig:failed}). One potential solution is to use multiple projectors covering each other's blind spots. In fact, extending the end-to-end full compensation framework to multiple projectors is an interesting future direction.

	{\small
		\bibliographystyle{ieee_fullname}
		\bibliography{ref}
	}
	
\end{document}